\documentclass{article} % For LaTeX2e
\usepackage{iclr2027_conference,times}

\usepackage{hyperref}
\usepackage{url}
\usepackage{booktabs}
\usepackage[table]{xcolor}
\usepackage{makecell}
\usepackage{graphicx}
\usepackage{subcaption}
\usepackage{tcolorbox}
\usepackage{xcolor}
\usepackage{newtxtext,newtxmath}  % 或 fontspec + \setmainfont{Times New Roman}

\newtcolorbox{formatbox}[1][]{
    colback=gray!10,
    colframe=gray!40!black,
    colbacktitle=gray!40!black,
    coltitle=white,
    fonttitle=\rmfamily,
    fontupper=\rmfamily,
    arc=3mm,
    boxrule=0.5mm,
    left=4mm, right=4mm, top=2mm, bottom=2mm,
    #1
}

\title{EG-GRPO: Evidence-Guided Step-level Credit \\ Assignment for Deep Search Agents}

\author{Antiquus S.~Hippocampus, Natalia Cerebro \& Amelie P. Amygdale \thanks{ Use footnote for providing further information
about author (webpage, alternative address)---\emph{not} for acknowledging
funding agencies.  Funding acknowledgements go at the end of the paper.} \\
Department of Computer Science\\
Cranberry-Lemon University\\
Pittsburgh, PA 15213, USA \\
\texttt{\{hippo,brain,jen\}@cs.cranberry-lemon.edu} \\
\And
Ji Q. Ren \& Yevgeny LeNet \\
Department of Computational Neuroscience \\
University of the Witwatersrand \\
Joburg, South Africa \\
\texttt{\{robot,net\}@wits.ac.za} \\
\AND
Coauthor \\
Affiliation \\
Address \\
\texttt{email}
}

\iclrfinalcopy % Uncomment for camera-ready version, but NOT for submission.

\author{
\textbf{Sijia Luo\(^{1,2,5}\)}, \textbf{Mengjie Gao\(^{5}\)}, \textbf{Yuxiao He\(^{5}\)}, \textbf{Mengshu Sun\(^{5}\)},\\ \textbf{Cuiping Li\(^{1,4}\)}, \textbf{Hong Chen\(^{1,3}\)}, \textbf{Jing Zhang\(^{1,2}\)\thanks{Corresponding Author.}} \\
\(^1\) Renmin University of China, Beijing 100872, China \\
\(^2\) Key Laboratory of Data Engineering and Knowledge Engineering, MOE, China \\
\(^3\) Engineering Research Center of Database and Business Intelligence, MOE, China \\
\(^4\) Beijing Key Laboratory of Intelligent Database Systems and Applications, China \\
\(^5\) Ant Group, Beijing, China \\
\texttt{sijialuo23@gmail.com} \\
\texttt{\{liuxiao.gmj, matthew.hyx, mengshu.sms\}@antgroup.com} \\
\texttt{\{licuiping, chong, zhang-jing\}@ruc.edu.cn}
}

\begin{document}

\maketitle

\begin{abstract}
% LLM-based search agents demonstrate strong long-horizon retrieval capabilities, typically trained via Reinforcement Learning using outcome rewards. However, this trajectory-level supervision introduces severe reward sparsity, failing to provide precise credit assignment for intermediate multi-turn search actions. While process supervision can alleviate this, defining reliable step-level signals for open-ended search remains challenging. In this paper, we propose Evidence Graph-guided Group Relative Policy Optimization (EG-GRPO), a novel framework that leverages the evidence graphs inherently generated within synthetic training data to provide task-grounded, step-level credit assignment. By tracking the progressive discovery of required factual claims, EG-GRPO formulates dense process rewards without requiring manual annotation or auxiliary reward models. To ensure robust policy optimization, we introduce a cumulative knowledge state representation and a hierarchical fallback normalization strategy, enabling fair advantage estimation across divergent search trajectories. Extensive experiments across five complex deep search benchmarks reveal that EG-GRPO consistently outperforms standard GRPO and advanced fine-grained baselines at both 4B and 30B scales. Our analysis demonstrates that EG-GRPO effectively mitigates premature search termination, guiding agents to sustain exploration and systematically acquire necessary evidence.

LLM-based search agents demonstrate strong capabilities in long-horizon information seeking and reasoning, typically trained via reinforcement learning. However, existing approaches predominantly rely on trajectory-level outcome rewards, providing limited supervision for complex search and reasoning processes. Although step-level credit assignment offers a promising solution, obtaining reliable and task-grounded supervision for open-ended search remains challenging. In this paper, we propose \textbf{E}vidence-\textbf{G}uided \textbf{G}roup \textbf{R}elative \textbf{P}olicy \textbf{O}ptimization (\textbf{EG-GRPO}), a framework that derives step-level supervision from the evidence annotations available in synthetic deep search data. EG-GRPO identifies the first supporting citation for each required evidence claim and uses the newly acquired evidence to construct step-level process rewards. By representing the accumulated evidence as an evolving knowledge state, EG-GRPO further performs state-conditioned relative advantage estimation and combines the resulting step-level signal with the standard trajectory-level outcome advantage. Experiments on five challenging deep search benchmarks show that EG-GRPO generally outperforms standard GRPO and existing reward-enhanced and step-level RL baselines across different model scales. These results demonstrate that task-grounded evidence annotations can provide effective process supervision for credit assignment in deep search RL.

\end{abstract}

\section{Introduction}
\label{sec:intro}
As the capabilities of Large Language Models (LLMs) continue to advance, deep search agents have demonstrated immense potential in tackling knowledge-intensive tasks~\citep{openai2025deepresearch, xi2025surveyllmbaseddeepsearch, Wang_2024}. Grounded in the ReAct paradigm, these agents iteratively gather evidence within noisy web environments by alternating between reasoning, tool utilization, and observation, ultimately providing well-supported answers~\citep{yao2023reactsynergizingreasoningacting}. To further enhance their long-horizon information retrieval capabilities, Reinforcement Learning (RL) has emerged as the prevalent paradigm for training deep search agents. Existing work typically uses a binary outcome reward as the training signal on synthetic multi‑hop QA datasets, comparing the agent’s final answer against the ground truth~\citep{jin2025searchr1trainingllmsreason, gao2025turnsunlockinglonghorizonagentic, li2025websailornavigatingsuperhumanreasoning, liu2025webexplorerexploreevolvetraining}.
% among which Group Relative Policy Optimization (GRPO) and its variants are most widely adopted. By comparing the agent's final answer against the ground truth and utilizing a binary outcome reward as the training signal, these approaches have achieved remarkable progress across multiple deep search benchmarks.

In deep search scenarios, a complete trajectory typically comprises anywhere from several to over a dozen retrieval iterations, with each retrieval action potentially exerting a critical impact on the final answer formulation. However, the entire trajectory receives only a single binary scalar signal at its conclusion, leaving the model unable to discern which specific retrieval step played the decisive role~\citep{shao2025deepseekmath}. To alleviate this reward sparsity issue, several studies have attempted to introduce richer reward signals at the trajectory level~\citep{gunjal2026rubrics, lu2025deepdiveadvancingdeepsearch}. E-GRPO~\citep{zhao2026repurposingsyntheticdatafinegrained} leverages entity annotations in synthetic data to formulate an entity matching rate as a fine-grained reward; CaRR~\citep{zhang2026chainingevidencerobustreinforcement} further decomposes questions into single-hop rubrics, combining context-aware rubric rewards and outcome rewards in RL. Although these methods yield substantial improvements in their respective evaluation settings, they fundamentally still calculate a scalar reward for the entire trajectory. Consequently, the core issue of differentiated step-level credit assignment remains unresolved.

Step-level credit assignment provides a more direct approach to alleviating the aforementioned problem and has been extensively studied in sequential reasoning~\citep{zhang2026reasoning, guo2026segment, wang2026information}. Process Reward Models (PRMs) provide step-level supervision over intermediate reasoning steps, offering a more direct training signal than outcome-only rewards~\citep{lightman2024let, pala2025error}. Meanwhile, GiGPO~\citep{feng2025groupingrouppolicyoptimizationllm} introduces state-based grouping to estimate step-level relative advantages for long-horizon LLM agents. Recently, STAMP~\citep{xu2026stampprovenanceguidedcreditassignment} further extends step-level credit assignment to deep search agents by tracing evidence provenance and attributing credit to the first retrieval action that exposes supporting evidence. 
However, applying such approaches to open-ended deep search still faces two key challenges. First, search actions lack a well-defined notion of local correctness: a retrieval step may be valuable for discovering direct evidence, uncovering indirect clues, or narrowing the search space, making reliable and task-grounded step-level supervision difficult to obtain. Second, fine-grained credit assignment requires a meaningful representation of the evolving search state, whereas raw multi-turn search histories are highly heterogeneous and difficult to match reliably. These challenges motivate us to look beyond external process supervision and ask whether a task-grounded step-level signal is already available in the synthetic data used to train deep search agents.
% However, applying step-level credit assignment to open-ended deep search remains challenging. In particular, search trajectories lack a well-defined notion of local correctness, making it difficult to obtain reliable and task-grounded supervision for intermediate retrieval actions. Moreover, the evolving information state of a multi-turn search trajectory is difficult to represent and compare reliably. These challenges motivate us to explore whether task-grounded step-level supervision can be derived directly from the structured annotations already available in synthetic deep search data.

\begin{figure}[t]
  \centering
  \includegraphics[width=0.9\textwidth]{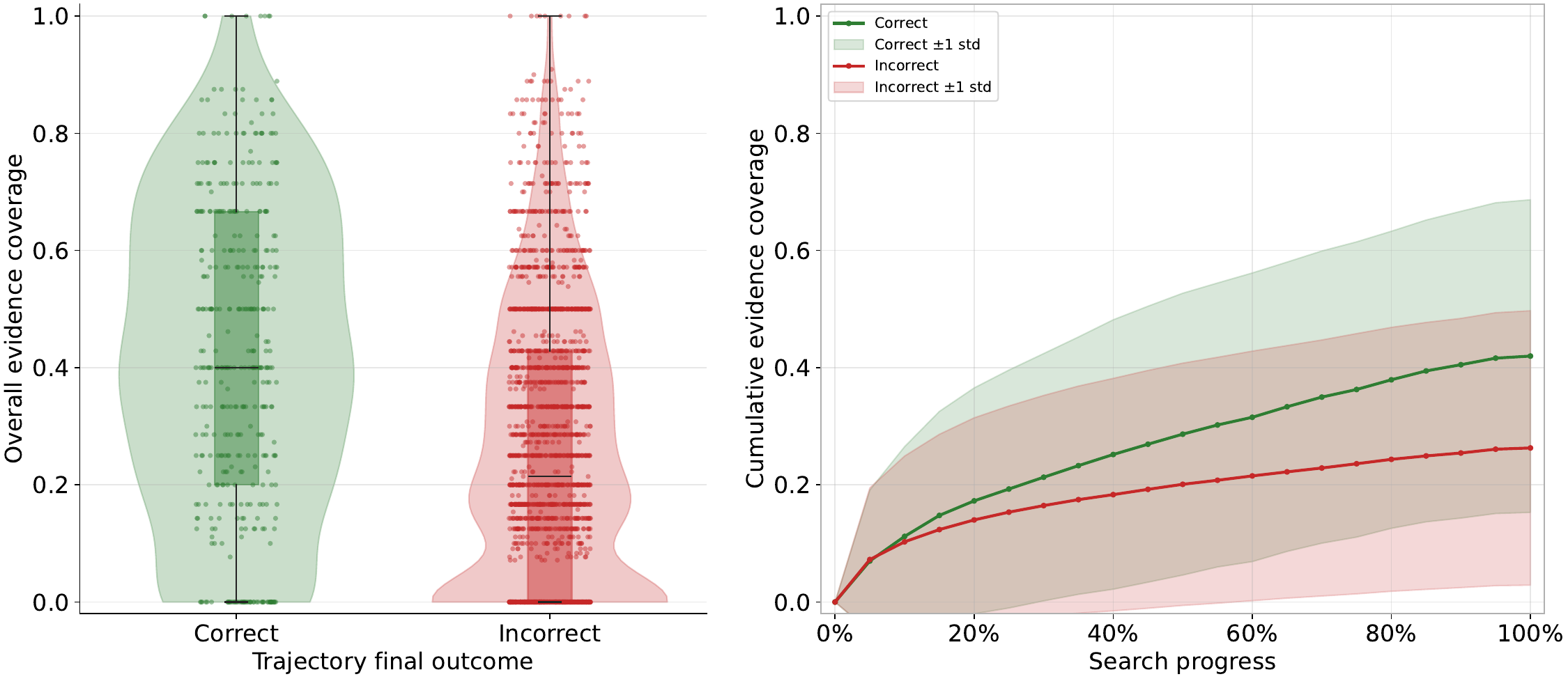}
  \caption{\textbf{Left:} Distribution of overall evidence coverage across correct and incorrect trajectories, showing that task success strongly correlates with higher coverage.
\textbf{Right:} Trajectory-wise cumulative evidence coverage over normalized search progress, demonstrating that correct trajectories consistently accumulate evidence faster from the early search stages.}
 \label{fig:pre-exp}
\end{figure}

We observe that such supervisory signals are inherently embedded in synthetic data pipelines. Current deep search datasets are often constructed through knowledge‑graph random walks and information obfuscation, which produce structured annotations describing the factual evidence needed to answer each query~\citep{lu2025deepdiveadvancingdeepsearch, du2026openseekerdemocratizingfrontiersearch}. Traditionally, only question‑answer pairs are retained for reward computation, discarding this rich annotation. However, the annotated evidence claims provide a natural source of task-grounded step-level supervision, as they specify the factual information that a successful trajectory should progressively acquire. To test this hypothesis, we conducted an experiment on a subset of the training set (320 prompts) by sampling 8 trajectories per prompt and computing the evidence coverage rate for each trajectory. Figure \ref{fig:pre-exp} (left) shows that correct trajectories have significantly higher coverage (median 41.97\% vs. 26.29\%), confirming coverage as a strong predictor of final correctness. Figure \ref{fig:pre-exp} (right) further reveals that correct trajectories maintain higher cumulative coverage from the very beginning, and this gap widens as retrieval proceeds. These observations indicate that evidence acquisition is a continuous process signal highly correlated with task success. More importantly, the cumulative set of acquired evidence claims provides an explicit representation of the agent's evolving knowledge state, allowing rollouts to be compared under identical evidence states without manual process annotation. This observation motivates us to use task-grounded evidence annotations as a source of step-level supervision for deep search RL.

Building upon these insights, we propose \textbf{E}vidence-\textbf{G}uided \textbf{G}roup \textbf{R}elative \textbf{P}olicy \textbf{O}ptimization (\textbf{EG-GRPO}), a step-level credit assignment method for deep search agents. EG-GRPO transforms the evidence claims annotated in synthetic data into dense process rewards and combines them with GRPO's trajectory-level advantages, preserving outcome supervision while providing fine-grained guidance for intermediate retrieval actions. Specifically, an LLM-based judge identifies the first supporting citation for each required evidence claim, and newly supported claims at each step are converted into process rewards. We further define a cumulative knowledge state based on the evidence claims acquired before each step and perform state-conditioned advantage normalization, enabling different rollouts to be compared under the same knowledge state. The resulting step-level advantages are then fused with trajectory-level advantages to produce token-level training signals. In this way, EG-GRPO provides task-grounded supervision for intermediate evidence acquisition while retaining the final outcome signal, enabling more precise credit assignment for deep search agents.

We systematically evaluate \textbf{EG-GRPO} on five deep search benchmarks: BrowseComp~\citep{wei2025browsecompsimplechallengingbenchmark}, BrowseComp-ZH~\citep{zhou2025browsecompzhbenchmarkingwebbrowsing}, GAIA~\citep{mialon2023gaiabenchmarkgeneralai}, xBenchDS-2505 and xBenchDS-2510~\citep{chen2025xbenchtrackingagentsproductivity}. EG-GRPO consistently outperforms the GRPO baseline as well as reward-enhanced and step-level baselines, demonstrating that step-level credit assignment provides a complementary training signal to trajectory-level rewards. Our main contributions are summarized as follows:

\begin{itemize}
\item Through step-level analysis of rollout trajectories, we reveal a strong correlation between the quality of incremental evidence acquisition and final answer correctness, with the most pronounced divergence emerging early in the search process. This provides empirical evidence for the need for step-level credit assignment in deep search RL.

\item We propose \textbf{EG-GRPO}, which transforms task-grounded evidence claims from synthetic training data into dense step-level rewards and employs hierarchical fallback normalization for stable advantage estimation, without requiring additional human annotations or auxiliary model training.

\item We conduct extensive experiments across five deep search benchmarks, demonstrating the effectiveness of EG-GRPO. Comprehensive ablation studies further validate the contribution of its key components.

\end{itemize}

\section{Preliminary}
\label{sec:preliminary}
\subsection{Deep Search Agents}
\label{subsec:search_agents}

% \paragraph{Multi-Turn Rollout.}
% We adopt the ReAct paradigm~\citep{yao2023reactsynergizingreasoningacting} for deep search agents. Given a query $q$, an LLM agent policy $\pi_\theta$ iteratively interacts with an external environment (e.g., web search engines and browser tools) to gather factual evidence and derive the final answer. 
% At each step $t \in \{1, 2, \dots, T\}$, the agent generates a textual thought trace $\tau_t$ and an action $a_t \in \mathcal{A}$ (e.g., executing a web search query or visiting a web page) based on the query $q$ and the interaction context up to step $t-1$:
% \begin{equation}
% (\tau_t, a_t) \sim \pi_\theta(\cdot \mid q, \mathcal{H}_{<t}),
% \end{equation}
% where $\mathcal{H}_{<t} = (\tau_1, a_1, o_1, \dots, \tau_{t-1}, a_{t-1}, o_{t-1})$ denotes the history context before step $t$. 
% The environment then returns an observation $o_t$ (e.g., retrieved search snippets or scraped page contents). 
% The multi-turn rollout terminates when the agent emits a special stopping signal and outputs the final predicted answer $y$. A full trajectory $\mathcal{H}$ of length $T$ is defined as:
% \begin{equation}
% \mathcal{H} = (\tau_1, a_1, o_1, \dots, \tau_T, a_T, o_T, y).
% \end{equation}

\paragraph{Multi-Turn Rollout.}
We adopt the ReAct paradigm~\citep{yao2023reactsynergizingreasoningacting} for deep search agents. Given a query $q$, an LLM agent policy $\pi_\theta$ iteratively interacts with an external environment, such as web search engines and browser tools, to gather factual evidence and derive the final answer. At each turn $t \in \{1,2,\dots,T\}$, the agent generates a textual reasoning trace $\tau_{i,t}$ and an action $a_{i,t}\in\mathcal{A}$ conditioned on the query and the interaction history:
\begin{equation}
(\tau_{i,t}, a_{i,t})
\sim
\pi_\theta(\cdot \mid q, \mathcal{H}_{i,<t}),
\end{equation}
where
\begin{equation}
\mathcal{H}_{i,<t}
=
(\tau_{i,1},a_{i,1},o_{i,1},\dots,
\tau_{i,t-1},a_{i,t-1},o_{i,t-1})
\end{equation}
denotes the interaction history before turn $t$, and $o_{i,t}$ is the environment observation returned after executing $a_{i,t}$, such as retrieved search snippets or web page contents.

The multi-turn rollout terminates when the agent emits a stopping signal and produces a final answer $y_i$. We denote the resulting trajectory by
\begin{equation}
\tau_i =
(\tau_{i,1},a_{i,1},o_{i,1},\dots,
\tau_{i,T},a_{i,T},o_{i,T},y_i).
\end{equation}

\subsection{Evidence Annotations in Synthetic Data}
\label{subsec:evidence_graph}

% \paragraph{Synthetic Multi-Hop Data Generation.}
% Modern deep search datasets~\citep{lu2025deepdiveadvancingdeepsearch, du2026openseekerdemocratizingfrontiersearch, chu2026redsearcherscalablecostefficientframework} rely on entity-centric knowledge graph traversal and information obfuscation to construct complex, multi-hop questions automatically. During this generation process, a structured factual path is traversed to connect a seed entity to the target answer, generating a rich underlying dependency graph.

% \paragraph{Evidence Graph Formalization.}
% We formalize this structured artifact associated with query $q$ as an evidence graph $\mathcal{G} = (\mathcal{V}, \mathcal{E})$. The node set $\mathcal{V} = \{c_1, c_2, \dots, c_M\}$ comprises $M$ atomic evidence claims (or factual statements) required to fully resolve query $q$. Each claim $c_m$ represents an indivisible factual assertion. The directed edge set $\mathcal{E} \subseteq \mathcal{V} \times \mathcal{V}$ models the relational structure among claims. A directed edge $(c_m, c_n) \in \mathcal{E}$ means that claim $c_m$ is directly connected to claim $c_n$ in the relational graph.

\paragraph{Synthetic Multi-Hop Data Generation.}
Modern deep search datasets~\citep{lu2025deepdiveadvancingdeepsearch, du2026openseekerdemocratizingfrontiersearch, chu2026redsearcherscalablecostefficientframework} leverage entity-centric knowledge graph traversal and information obfuscation to automatically construct complex multi-hop questions. During data generation, a structured sequence of factual relations is used to connect a seed entity to the target answer, yielding a set of intermediate factual claims required to resolve the query.

\paragraph{Evidence Claim Formalization.}
For each query $q$, we denote the annotated evidence claims produced during synthetic data generation as
\begin{equation}
\mathcal{C}_q = \{c_1,\ldots,c_K\},
\end{equation}
% where $\mathcal{C}_q=\{c_1,\dots,c_K\}$ is the set of $K$ atomic evidence claims required to resolve $q$, and $\mathcal{R}_q\subseteq\mathcal{C}_q\times\mathcal{C}_q$ represents the dependency relations among these claims. Each claim $c_k$ corresponds to an indivisible factual assertion. A directed edge $(c_j,c_k)\in\mathcal{R}_q$ indicates that claim $c_j$ provides a direct prerequisite or dependency for establishing claim $c_k$.

% The evidence graph is generated offline together with the synthetic training data. In EG-GRPO, we use the claim set $\mathcal{C}_q$ to derive step-level evidence rewards.
where $\mathcal{C}_q$ contains $K$ atomic evidence claims required to resolve $q$. Each claim $c_k$ corresponds to an indivisible factual assertion. These claims are generated offline together with the synthetic training data and provide task-grounded supervision for evidence acquisition. In EG-GRPO, we use $\mathcal{C}_q$ to derive step-level evidence rewards and to construct the evolving knowledge state.

\section{Methodology}
\label{sec:method}

% EG-GRPO addresses the step-level credit assignment gap in outcome-supervised RL for deep search agents by converting an offline evidence graph into dense step-level rewards, which are then fused with trajectory-level advantages.

% Formally, each rollout $\tau_i$ produces two reward signals:
% \begin{itemize}
%     \item \textbf{Outcome reward} $R_i^{\text{out}} \in \{0, 1\}$: a binary signal indicating whether the agent's final answer is correct .
%     \item \textbf{Step rewards} $\{r_{i,t}\}_{t=1}^{T_i}$: a sequence of per-step scalars measuring the fraction of newly retrieved evidence claims at each retrieval step (detailed in \S\ref{sec:step_reward}).
% \end{itemize}

% The two signals give rise to two advantage streams—a trajectory-level advantage $A_i^{\text{traj}}$ from GRPO and a step-level advantage $A_{i,t}^{\text{step}}$ from discounted return normalization—which are fused into a per-token advantage:
% \begin{equation}
%     \hat{A}_{i,t} = A_i^{\text{traj}} + 
%     w \cdot \mathrm{clip}\!\left(A_{i,t}^{\text{step}},\,-C,\,C\right),
%     \label{eq:total_adv}
% \end{equation}
% where $w = 0.5$ is a fixed fusion weight and $C$ is a clipping bound that prevents any single step from dominating the gradient signal.

\subsection{Evidence Claims and Step-Level Rewards}
\label{sec:step_reward}

% \paragraph{Evidence graph construction.}
% For each training query $q$, we construct an evidence graph $\mathcal{G}_q=(\mathcal{C}_q,\mathcal{R}_q)$ offline alongside the synthetic data generation pipeline. The graph contains a set of atomic evidence claims
% $\mathcal{C}_q=\{c_1,\dots,c_K\}$ that collectively form the factual backbone required to answer $q$. Each claim is assigned a uniform weight $1/K$. The evidence graph is used only during training for step-level reward computation.
\paragraph{Evidence Claim Construction.}
For each training query $q$, we obtain a set of atomic evidence claims
\begin{equation}
\mathcal{C}_q=\{c_1,\ldots,c_K\}
\end{equation}
from the synthetic data generation pipeline. These claims collectively specify the factual evidence required to resolve $q$. Each claim is assigned a uniform weight $1/K$. The evidence claims are used during training to provide task-grounded supervision for step-level evidence acquisition.

\paragraph{Evidence-to-trajectory attribution.}
Given a completed rollout $\tau_i$, we first compress the tool responses into a set of numbered citation blocks. A single LLM judge reads the evidence claims $\mathcal{C}_q$ together with the citation contents and identifies, for each claim $c_k$, the earliest citation that explicitly supports it. By mapping each citation to its originating message and then to its corresponding turn, we obtain the first-exposure function:
\begin{equation}
\phi_i(c_k)
=
\min\{t : c_k \text{ is first supported at turn } t\}.
\end{equation}
Each claim is therefore credited only at its earliest exposure turn, preventing repeated retrieval of the same evidence from receiving additional reward.

\begin{figure}[t]
  \centering
  \includegraphics[width=1.0\textwidth]{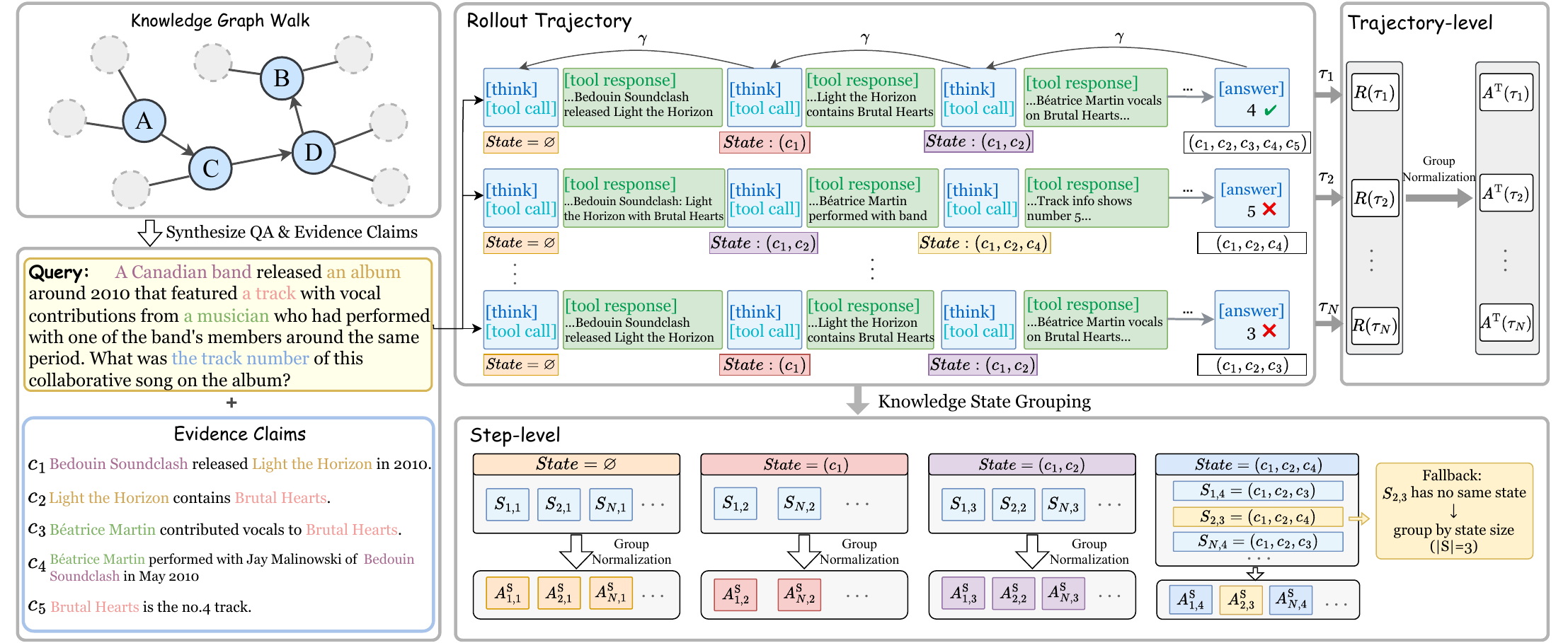}
  \caption{Overview of EG-GRPO. Step-level advantages ($A^S$) are computed through fine-grained knowledge-state grouping. Intermediate steps across different trajectories are aligned according to their cumulative knowledge states. A three-level fallback mechanism progressively relaxes the grouping criterion when exact states are sparse, enabling robust step-level advantage estimation.}
  \label{fig:overview}
\end{figure}

% \paragraph{Incremental rewards and cumulative knowledge state.}
% Drawing on the definition of anchor state from GiGPO~\citep{feng2025groupingrouppolicyoptimizationllm}, we define the \emph{cumulative knowledge state} at step $t$ as the set of evidence claims already discovered before turn $t$:
% \begin{equation}
%     S_{i,t} = \{c_k : \phi(c_k) < t\}.
% \end{equation}
% The newly covered evidence at step $t$ is $\Delta_{i,t} = \{c_k : \phi(c_k) = t\}$, and the instantaneous step reward is:
% \begin{equation}
%     r_{i,t} = \frac{|\Delta_{i,t}|}{K} \in \left[0,\, \frac{1}{K}\right].
% \end{equation}
% The trajectory-level evidence coverage $R_i^{\text{cov}} = \sum_t r_{i,t} = |\bigcup_t \Delta_{i,t}| / K$ is retained as a monitoring quantity but does not enter the training objective directly.

% A key property of $S_{i,t}$ is that it represents the agent's \emph{cognitive state} upon entering step $t$: two rollouts on the same prompt that share an identical $S_{i,t}$ are in a directly comparable epistemic position, providing the basis for meaningful state grouping (see \S\ref{sec:step_adv}).

% \paragraph{Special cases.}
% If the final response is malformed or absent, we set $r_{i,T} = -1$ as a terminal penalty. If the answer is correctly produced with no further tool calls, the final step is marked with 
% a sentinel state $S_{i,T} = \{\texttt{SUMMARY\_STEP}\}$. Rollouts that time out or encounter tool failures are excluded from policy updates.

\paragraph{Incremental rewards and cumulative knowledge state.}
For each rollout $\tau_i$, we define the \emph{cumulative knowledge state} before turn $t$ as the set of evidence claims already discovered:
\begin{equation}
S_{i,t}
=
\{c_k\in\mathcal{C}_q:\phi_i(c_k)<t\}.
\end{equation}
The newly discovered evidence at turn $t$ is
\begin{equation}
\Delta_{i,t}
=
\{c_k\in\mathcal{C}_q:\phi_i(c_k)=t\},
\end{equation}
and the corresponding step reward is
\begin{equation}
r_{i,t}
=
\frac{|\Delta_{i,t}|}{K}
\in
\left[0,1\right].
\end{equation}

The trajectory-level evidence coverage is defined as
\begin{equation}
R_i^{\mathrm{cov}}
=
\sum_{t=1}^{T} r_{i,t}
=
\frac{\left|\bigcup_{t=1}^{T}\Delta_{i,t}\right|}{K}.
\end{equation}

Importantly, $S_{i,t}$ summarizes the evidence acquired by the agent before taking the action at turn $t$. Thus, two rollouts for the same query that share the same $S_{i,t}$ are aligned at the same knowledge state, providing a meaningful basis for state-conditioned advantage normalization (Section~\ref{sec:step_adv}).

\paragraph{Special cases.}
If the final response is malformed or absent, we assign a terminal penalty $r_{i,T}=-1$. If the agent produces the final answer correctly without any further tool call, the final action is marked with a sentinel state $S_{i,T}=\{\texttt{SUMMARY\_STEP}\}$. Rollouts that time out or encounter unrecoverable tool failures are excluded from policy updates.

\subsection{Trajectory-Level Advantage: GRPO}

% We group all $N$ rollouts sharing the same prompt uid and apply standard GRPO group normalization to the outcome rewards:
% \begin{equation}
% A^T(\tau_i) = \frac{R(\tau_i) - \text{mean}(\{R(\tau_j)\}_{j=1}^N)}{F_{\text{norm}}(\{R(\tau_j)\}_{j=1}^N)}.
% \end{equation}
% The trajectory-level advantage $A^T(\tau_i)$ is then broadcast uniformly to every action token in rollout $\tau_i$.

For each query $q$, we sample $N$ rollouts
$\{\tau_i\}_{i=1}^{N}$ and apply standard GRPO group normalization to their outcome rewards. Denoting the outcome reward of trajectory $\tau_i$ by $R_i^{\mathrm{out}}$, the trajectory-level advantage is
\begin{equation}
A_i^T
=
\frac{
R_i^{\mathrm{out}}
-
\operatorname{mean}_{j=1}^{N}
R_j^{\mathrm{out}}
}{
\operatorname{std}(\{R_j^{\mathrm{out}}\}_{j=1}^{N})
}.
\end{equation}
The trajectory-level advantage $A_i^T$ is broadcast to all decision-related tokens in trajectory $\tau_i$, following the standard GRPO formulation.

\subsection{Step-Level Advantage: Discounted Return and 
Three-Level Fallback Normalization}
\label{sec:step_adv}

% \paragraph{Discounted step return.}
% To better capture long-horizon impact, we compute the discounted cumulative step reward from turn $t$:
% \begin{equation}
%     R_t^{(i)} = \sum_{k=t}^{T} \gamma^{k-t}\, r_k^{(i)},
% \end{equation}
% with discount factor $\gamma$. When the terminal step carries a format-error penalty ($r_T^{(i)} < 0$), its contribution is retained at the final step but is not propagated backward through $\gamma$, preventing the negative penalty from contaminating the return estimates of earlier evidence-producing steps.

\paragraph{Discounted step return.}
To capture the delayed impact of evidence acquisition, we compute the discounted return from turn $t$:
\begin{equation}
G_{i,t}
=
\sum_{k=t}^{T}
\gamma^{k-t}r_{i,k},
\end{equation}
where $\gamma$ is the discount factor. When the terminal step carries a format-error penalty ($r_{i,T}<0$), this penalty is retained at the terminal step but is not propagated backward through $\gamma$, preventing a formatting failure from contaminating the credit assigned to earlier evidence-acquisition steps.

\paragraph{Three-level fallback grouping.}
A naive strategy would normalize $G_{i,t}$ across all steps associated with the same query. However, steps taken under different knowledge states are not directly comparable: an action taken before any evidence has been acquired faces a different retrieval context from one taken after several claims have already been established.

We therefore normalize step returns using the finest available knowledge-state grouping and progressively relax the grouping criterion when the finer group contains insufficient samples. Specifically, we use the following three levels:

\begin{table}[h]
\centering
\small
\begin{tabular}{lcc}
\toprule
\textbf{Level} & \textbf{Grouping key} & \textbf{Trigger condition} \\
\midrule
1 (exact state)  & $(q,\;S_{i,t})$ & $\ge M_1$ samples \\
2 (state size)   & $(q,\;|S_{i,t}|)$ & $\ge M_1$ samples \\
3 (query global) & $q$ & $\ge 2$ samples \\
\bottomrule
\end{tabular}
\caption{Three-level fallback grouping for step-level advantage estimation.}
\label{tab:fallback}
\end{table}

Let $\mathcal{G}_{i,t}$ denote the first grouping level that satisfies its corresponding trigger condition for step $(i,t)$. The step-level advantage is then computed by normalizing its discounted return within the selected group:
\begin{equation}
A_{i,t}^{S}
=
\frac{
G_{i,t}
-
\operatorname{mean}
\left(
\{G_{j,u}:(j,u)\in\mathcal{G}_{i,t}\}
\right)
}{
\operatorname{std}
\left(
\{G_{j,u}:(j,u)\in\mathcal{G}_{i,t}\}
\right)
}.
\end{equation}
If the exact-state group contains insufficient samples, we fall back to the state-size group; if that group is also insufficient, we use the query-global group. The final level requires only two samples within the query group, ensuring that a normalization baseline is available for valid rollouts.

The hierarchy provides a trade-off between precision and statistical robustness. Level 1 compares steps taken from identical knowledge states and therefore provides the most fine-grained credit signal. However, exact states can be sparse because different rollouts often acquire evidence in different orders. Level 2 relaxes the criterion to states containing the same number of known claims, preserving a coarse notion of comparability while increasing the number of matched samples. Level 3 further falls back to all steps associated with the same query, ensuring broad coverage when finer-grained matches are unavailable.

After normalization, the step-level advantage is clipped to $[-C,C]$. The resulting scalar is broadcast to all decision-related tokens within the corresponding turn segment, providing a dense and forward-looking supervisory signal for policy optimization.

\subsection{Training Objective}

% The two signals give rise to two advantage streams—a trajectory-level advantage $A^T(\tau_i)$ from GRPO and a step-level advantage $A^S(\boldsymbol{a}_t^{(i)})$ from discounted return normalization—which are fused into a per-token advantage:
% \begin{equation}
%     A(\boldsymbol{a}_t^{(i)}) = A^T(\tau_i) + 
%     w \cdot \mathrm{clip}\!\left(A^S(\boldsymbol{a}_t^{(i)}),\,1\pm C\right),
%     \label{eq:total_adv}
% \end{equation}
% where $w$ is a fixed weight and $C$ is a clipping bound that prevents any single step from dominating the gradient signal.

% The policy is updated by maximizing a clipped surrogate objective over all action tokens:

% \begin{equation}
% \label{eq:final_objective}
% \begin{aligned}
% \mathcal{J}_{\text{EG-GRPO}}(\theta) =  \mathbb{E}_{\substack{q \sim \mathcal{D} \\ \tau_i \sim \pi_{\theta_{\text{old}}}}} 
% \Bigg[ \frac{1}{NT} \sum_{i=1}^{N} \sum_{t=1}^{T} \min \Big( \rho_\theta(\boldsymbol{a}_t^{(i)})) A(\boldsymbol{a}_t^{(i)}),  \text{clip}\big(\rho_\theta(\boldsymbol{a}_t^{(i)}), 1\pm\epsilon\big) A(\boldsymbol{a}_t^{(i)}) \Big)  \Bigg]
% \end{aligned}
% \end{equation}

% where $\rho_{\theta}(a_t^{(i)}) = \frac{\pi_{\theta}(a_t^{(i)} | s_t^{(i)}, q)}{\pi_{\theta_{old}}(a_t^{(i)} | s_t^{(i)}, q)}$ is the importance sampling ratio. The optimization procedure restricts gradient propagation to decision-related tokens (reasoning, tool calls, and answers), while masking the raw tool responses.

The two advantage streams provide complementary supervision: the trajectory-level advantage $A_i^T$ captures whether the complete rollout successfully solves the query, while the step-level advantage $A_{i,t}^{S}$ provides finer-grained credit for evidence acquisition. We combine them into a per-token advantage:
\begin{equation}
A_{i,t}
=
A_i^T
+
w\cdot
\operatorname{clip}
\left(
A_{i,t}^{S},-C,C
\right),
\label{eq:total_adv}
\end{equation}
where $w$ controls the contribution of step-level supervision and $C$ bounds the magnitude of the step-level signal.

The policy is optimized using a clipped surrogate objective:
\begin{equation}
\label{eq:final_objective}
\begin{aligned}
\mathcal{J}_{\mathrm{EG\text{-}GRPO}}(\theta)
=
\mathbb{E}
\Bigg[
\frac{1}{N}
\sum_{i=1}^{N}
\frac{1}{|\mathcal{T}_i|}
\sum_{t\in\mathcal{T}_i}
\min
\Big(
\rho_{i,t}(\theta)A_{i,t},
\operatorname{clip}
\big(
\rho_{i,t}(\theta),1-\epsilon,1+\epsilon
\big)
A_{i,t}
\Big)
\Bigg],
\end{aligned}
\end{equation}
where $\mathcal{T}_i$ denotes the decision-token positions in trajectory $\tau_i$, and
\begin{equation}
\rho_{i,t}(\theta)
=
\frac{
\pi_\theta(x_{i,t}\mid x_{i,<t},q)
}{
\pi_{\theta_{\mathrm{old}}}(x_{i,t}\mid x_{i,<t},q)
}
\end{equation}
is the token-level importance sampling ratio. Here, $x_{i,t}$ denotes a decision-related token, including reasoning, tool-call, or answer tokens.

The optimization masks out raw tool-response tokens, so gradients are propagated only through the agent's decision-related tokens. The same turn-level step advantage $A_{i,t}^{S}$ is assigned to all decision tokens within the corresponding turn segment.

\section{Experiments}
\label{sec:exp}
\subsection{Experiment Setup}
\begin{table}[!ht]
\centering
\caption{Overall performance of different agents on five challenging deep search benchmarks.}
\label{tab:main_result}
\begin{tabular}{@{}
    >{\raggedright}p{0.25\textwidth}   % Model
    >{\centering}p{0.12\textwidth}   
    >{\centering}p{0.12\textwidth}  
    >{\centering\arraybackslash}p{0.11\textwidth}  
    >{\centering\arraybackslash}p{0.12\textwidth}  
    >{\centering\arraybackslash}p{0.12\textwidth}   
    @{}}
\toprule
\textbf{Model} & \textbf{\makecell{ Browse \\ Comp }} & \textbf{\makecell{ Browse \\ Comp-ZH}} & \textbf{GAIA} & \textbf{\makecell{xBench- \\ DS-2505}} & \textbf{\makecell{xBench- \\ DS-2510}} \\
\midrule
\rowcolor{gray!20}
\multicolumn{6}{c}{\textbf{Advanced Agents with Proprietary Data}} \\
\midrule
GLM-4.7           & 67.5 & 66.6 & 61.9 & 72.0 & 52.3 \\
Kimi-K2.5         & 74.9 & 62.3 & -- & -- & 46.0   \\
DeepSeek-V3.2     & 67.6 & 65.0 & 75.1 & 78.0 & 55.7 \\
Gemini 3.1 Pro    & 53.6 & 75.9 & --   & 86.0 & --   \\
Tongyi-DeepResearch & 43.4 & 46.7 & 70.9 & 75.0 & --   \\
\midrule
\rowcolor{gray!20}
\multicolumn{6}{c}{\textbf{Agents with Open-source Data}} \\
\midrule
DeepDive-32B      & 14.8 & 25.6 & --   & 50.5 & --   \\
WebSailor-32B     & 10.5 & 25.5 & 53.2 & 53.3 & --   \\
WebExplorer-8B    & 15.2 & 32.0 & 50.0 & 53.7 & --   \\
OpenSeeker-30B    & 29.5 & 48.4 & --   & 74.0 & --   \\
REDSearcher-30B-A3B & 42.1 & 49.8 & --   & 80.1 & --   \\
\midrule
\rowcolor{gray!20}
\multicolumn{6}{c}{\textbf{Our Agents}} \\
\midrule
Qwen-4B-SFT       & 5.7  & 13.7 & 32.7 & 49.0  & 14.0  \\
+ GRPO             & 6.7  & 16.7 & 34.8 & 62.7 & 15.7 \\
+ C-GRPO           & 7.0   & 20.1 & \underline{38.2} & \underline{64.0}  & 13.7 \\
+ GiGPO            & 7.0   & 16.6 & 34.6 & 54.7 & 11.3  \\
% + IGPO              & 5.67 & 14.76 & 36.57 & 50.0  & 16.0   \\
+ STAMP             & \underline{7.3} & \underline{21.5} & 35.2 & 60.0  & \underline{17.0}   \\
+ EG-GRPO (ours)    & \textbf{8.0} & \textbf{23.5} & \textbf{45.9} & \textbf{65.0}  & \textbf{19.0}  \\
\midrule 
Qwen-30B-SFT      & 9.3  & 15.5 & 37.8 & 50.0  & 15.0  \\
+ GRPO             & 10.3  & 22.4 & 42.7 & 57.6 & 20.1  \\
+ C-GRPO           & 10.4  & 23.7 & \underline{47.3} & \underline{63.3} & 17.6 \\
+ GiGPO            & 10.5  & \underline{24.7} & 45.2 & 57.2 & 18.3 \\
% + IGPO              &       &       &       &       &       \\
+ STAMP             & \underline{10.8}      & 24.1      & 45.2      & 62.5      & \textbf{21.1}      \\
+ EG-GRPO (ours)    & \textbf{11.1}  & \textbf{26.1} & \textbf{49.1} & \textbf{64.2} & \underline{20.8} \\
\bottomrule
\end{tabular}
  \label{tab:main_result}
\end{table}

\paragraph{Tool server.}The agent engages with the live web through two tools: the \texttt{search} tool, which queries the Serper API~\citep{serper2025} and retrieves top‑10 snippets along with their associated URLs, and the \texttt{visit} tool, which extracts page content from a given URL via the Jina API~\citep{jina2025}.

\paragraph{Baselines.}
We compare EG-GRPO with the following RL algorithms: (1) GRPO~\citep{shao2024deepseekmathpushinglimitsmathematical}, which uses a binary outcome reward (0/1 score). (2) C-GRPO~\citep{zhang2026chainingevidencerobustreinforcement}, which combines a rubric-based citation score and outcome reward for training robust deep search agents. (3) GiGPO~\citep{feng2025groupingrouppolicyoptimizationllm}, which uses fine-grained credit assignment for LLM agents while preserving the appealing properties of group-based RL. (4) STAMP~\citep{xu2026stampprovenanceguidedcreditassignment}, which leverages provenance-guided supervision to perform fine-grained step-level credit assignment across search trajectories. Additionally, which leverages provenance-guided supervision to perform fine-grained step-level credit assignment across search trajectories. We also provide the reported scores of several state‑of‑the‑art search agents for reference (see Appendix \ref{app:reference} for details).

\paragraph{Training Details.}
% For the base models, we adopt two variants of the Qwen3 series: Qwen3-4B-Thinking-2507~\citep{yang2025qwen3} and Qwen3-30B-A3B-Thinking-2507~\citep{yang2025qwen3}, covering different model sizes and architectures(dense and MoE). The training pipeline consists of a cold-start SFT stage followed by RL training. For SFT, we select 1,000 high-quality trajectories from the OpenSeeker~\citep{du2026openseekerdemocratizingfrontiersearch} dataset and train each model for 1 epoch with a batch size of 32, a learning rate of 2e-5, and a maximum context length of 128k tokens. For RL, we use 2,316 synthesized training prompts. We set the per-prompt sampling number to 8, yielding a global batch size of 256 (i.e., $32$ prompts $\times 8$ samples) with a sampling temperature of 1.0. The RL models are trained for 2 epochs with a learning rate of 1e-6 and a maximum context length of 64k tokens. We employ Qwen-Flash~\citep{yang2025qwen3} as the judge LLM for both final-answer reward scoring and evidence verification. Full hyperparameters are provided in Appendix~\ref{app:training}.
We evaluate EG-GRPO across two base model architectures: Qwen3-4B-Thinking-2507 (dense) and Qwen3-30B-A3B-Thinking-2507 (MoE)~\citep{yang2025qwen3}. The training pipeline consists of a cold-start Supervised Fine-Tuning (SFT) stage using 1,000 high-quality trajectories selected from OpenSeeker~\citep{du2026openseekerdemocratizingfrontiersearch}, followed by RL training on 2,316 synthesized prompts. We employ Qwen-Flash~\citep{yang2025qwen3} as the judge LLM for both final-answer reward scoring and step-level evidence verification. Full hyperparameters, training parameters, and hardware configurations are detailed in Appendix~\ref{app:training}.

\paragraph{Benchmarks and Evaluation Details.}
The evaluation of our trained agents is conducted on five challenging deep search benchmarks: BrowseComp~\citep{wei2025browsecompsimplechallengingbenchmark}, BrowseComp‑ZH~\citep{zhou2025browsecompzhbenchmarkingwebbrowsing}, the text‑only validation subset of GAIA~\citep{mialon2023gaiabenchmarkgeneralai}, xBenchDS-2505 and xBenchDS-2510~\citep{chen2025xbenchtrackingagentsproductivity}. We use Qwen-Flash~\citep{yang2025qwen3} as judge LLM to determine if the agent’s final response agrees with the ground truth. We evaluate each model at 128k context length.

\subsection{Main Results}

\begin{figure}[htbp]
  \centering
  % 第一个子图
  \begin{subfigure}[b]{0.32\textwidth}
    \centering
    \caption{\textbf{Outcome Reward}}
    \label{fig:sub1}
    \includegraphics[width=\linewidth]{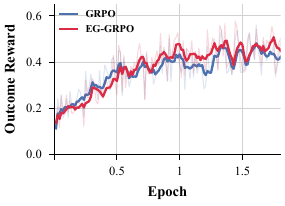}
  \end{subfigure}
  \hfill
  % 第二个子图
  \begin{subfigure}[b]{0.32\textwidth}
    \centering
    \caption{\textbf{Average Tool Call Steps}}
    \label{fig:sub2}
    \includegraphics[width=\linewidth]{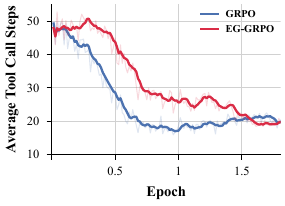}
  \end{subfigure}
  \hfill
  % 第三个子图
  \begin{subfigure}[b]{0.32\textwidth}
    \centering
    \caption{\textbf{Average Evidence Coverage}}
    \label{fig:sub3}
    \includegraphics[width=\linewidth]{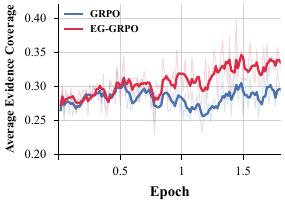}
  \end{subfigure}

  \caption{Training dynamics of GRPO and EG-GRPO on 4B models, including the changes of outcome rewards, average tool call steps and evidence coverage.}
  \label{fig:three_subplots}
\end{figure}

Table \ref{tab:main_result} summarizes the overall performance of EG-GRPO and various baselines across five challenging deep search benchmarks. Across both 4B and 30B model architectures, EG-GRPO generally outperforms the standard GRPO baseline as well as fine-grained reward-enhanced methods, achieving average accuracy gains of 4.96 and 3.64 over GRPO, respectively. Notably, GiGPO underperforms standard GRPO on both xBench-DS-2505 and xBench-DS-2510, suggesting that generic state-based grouping may be less effective for some long-horizon search tasks. In contrast, EG-GRPO demonstrates robust and consistent improvements; for instance, on the Qwen3-4B, it improves over standard GRPO by 11.11 absolute points on GAIA (45.9 vs. 34.8). These results demonstrate that task-grounded evidence supervision provides an effective complementary signal to trajectory-level outcome rewards for deep search RL.

\subsection{Analysis}

\paragraph{Training Dynamics.}
Figure \ref{fig:sub1} demonstrates that both GRPO and EG-GRPO successfully optimize the outcome reward during training, with EG-GRPO exhibiting comparable or slightly better performance in later epochs. Concurrently, Figure \ref{fig:sub2} reveals a distinct difference in the agents' search behaviors. While the baseline GRPO shows a sharp decline in average tool calls, suggesting a tendency to terminate searches prematurely, EG-GRPO maintains a more gradual reduction. This indicates that step-level supervision helps mitigate the risk of the agent taking superficial shortcuts, encouraging sufficient interaction with the environment before formulating an answer. 

\paragraph{Evidence Acquisition.}
The impact of this sustained exploration is reflected in the agents' evidence acquisition capabilities, as depicted in Figure \ref{fig:sub3}. While the standard GRPO model struggles to consistently increase its evidence coverage throughout training, EG-GRPO demonstrates a steady upward trend. This observation suggests that the evidence-guided step-level signals effectively guide the agent to progressively uncover the necessary factual evidence, rather than merely optimizing for the sparse outcome reward.

\subsection{Ablation Studies}

\paragraph{Weight of Step-Level Advantage.}

We investigate the balance between step-level evidence guidance and trajectory-level outcome supervision by varying the step advantage weight $w \in {0.0, 0.25, 0.5, 0.75, 1.0}$. As shown in Table \ref{tab:ab_weight}, introducing step-level guidance consistently improves performance over the $w=0$ GRPO baseline, with $w=0.5$ achieving the best results across all benchmarks. Notably, GAIA improves by 11.14\% absolute from $w=0$ to $w=0.5$. However, further increasing $w$ to 0.75 or 1.0 degrades performance, suggesting that excessive emphasis on local evidence acquisition can compromise global answer synthesis. Thus, $w=0.5$ provides the best balance between evidence gathering and final task completion.

\begin{table}[!ht]
\centering
\caption{Performance of EG-GRPO with different $w$.}
\label{tab:ab_weight}
\begin{tabular}{l | c c c c c}
\toprule
\textbf{Model} & \textbf{BC} & \textbf{BC-ZH} & \textbf{GAIA} & \textbf{xBenchDS-2505} & \textbf{xBenchDS-2510} \\
\midrule
Qwen3-4B-SFT        & 5.7  & 13.7 & 32.7 & 49.0  & 14.0  \\
+ EG-GRPO(w=0)       & 6.7  & 16.7 & 34.7 & 62.7 & 15.7 \\
+ EG-GRPO(w=0.25)    & 7.3  & 23.2 & 42.4 & 65.0  & 15.7 \\
+ EG-GRPO(w=0.5)     & \textbf{8.0} & \textbf{23.5} & \textbf{45.9} & \textbf{65.0}  & \textbf{19.0}  \\
+ EG-GRPO(w=0.75)    & 7.3  & 21.1 & 41.0 & 64.7 & 18.3 \\
+ EG-GRPO(w=1.0)     & 7.0   & 23.2 & 37.2 & 64.2 & 17.0  \\
\bottomrule
\end{tabular}
\end{table}

\paragraph{Effect of Fallback Grouping Strategy.}

We evaluate the three-level fallback normalization by comparing the full strategy with two boundary cases: Level 1 only (exact-state grouping) and Level 3 only (prompt-global normalization). As shown in Table \ref{tab:ab_fallback}, the full strategy achieves the best overall performance across the five benchmarks, although prompt-global normalization slightly outperforms it on GAIA. Exact-state grouping suffers from sparse matches among divergent rollouts, while global normalization ignores differences in intermediate knowledge states and introduces noisier comparisons. The hierarchical fallback strategy combines the benefits of both, providing fine-grained state conditioning when sufficient samples are available while maintaining robust normalization through progressive fallback.

\begin{table}[!ht]
\centering
\caption{Performance comparison of EG-GRPO under different fallback grouping strategies.}
\label{tab:ab_fallback}
\begin{tabular}{l | c c c c c}
\toprule
\textbf{Model} & \textbf{BC} & \textbf{BC-ZH} & \textbf{GAIA} & \textbf{xBenchDS-2505} & \textbf{xBenchDS-2510} \\
\midrule
EG-GRPO (full)          & \textbf{8.0} & \textbf{23.5} & 45.9 & \textbf{65.0} & \textbf{19.0} \\
EG-GRPO (level 3 only)  & 7.9  & 20.1 & \textbf{47.1} & 64.0 & 19.0 \\
EG-GRPO (level 1 only)  & 7.9  & 22.8 & 39.4 & 62.0 & 18.3 \\
\bottomrule
\end{tabular}
\end{table}

\section{Related Work}
\label{sec:related}
\paragraph{Reinforcement Learning for Deep Search Agents.}
Reinforcement learning has become an effective paradigm for training deep search agents to perform long-horizon information retrieval and reasoning. Early approaches such as Search-R1~\citep{jin2025searchr1trainingllmsreason} and DeepDive~\citep{lu2025deepdiveadvancingdeepsearch} employ outcome-based rewards on automatically constructed multi-hop search tasks, while subsequent systems further scale this paradigm through synthetic data generation and large-scale RL~\citep{gao2025turnsunlockinglonghorizonagentic, du2026openseekerdemocratizingfrontiersearch, chu2026redsearcherscalablecostefficientframework}. However, binary outcome rewards provide limited supervision for long search trajectories. Recent work therefore exploits auxiliary annotations from synthetic data to construct richer trajectory-level rewards. E-GRPO~\citep{zhao2026repurposingsyntheticdatafinegrained} leverages ground-truth entity information to derive dense entity-aware rewards, while CaRR~\citep{zhang2026chainingevidencerobustreinforcement} decomposes complex questions into verifiable single-hop rubrics and incorporates citation correctness into the reward. These approaches provide more informative supervision, but their rewards remain scalar signals assigned at the trajectory level. 
% In contrast, EG-GRPO further decomposes such synthetic evidence annotations into turn-level signals and explicitly assigns credit to individual evidence-acquisition steps.

\paragraph{Step-Level Credit Assignment.}

Process Reward Models (PRMs) provide intermediate-step supervision for mathematical reasoning~\citep{zhang2025lessonsdevelopingprocessreward}, while GiGPO~\citep{feng2025groupingrouppolicyoptimizationllm} introduces state-based grouping to estimate step-level relative advantages for long-horizon LLM agents. More recently, STAMP~\citep{xu2026stampprovenanceguidedcreditassignment} further assigns step-level credit through provenance tracing, identifying the first retrieval action that exposes supporting evidence. 
% In contrast, EG-GRPO derives task-grounded process supervision directly from the evidence claims available in synthetic training data. By tracking the first exposure of each required claim and representing the accumulated evidence as an explicit knowledge state, EG-GRPO enables state-conditioned step-level advantage estimation without additional human process annotations or auxiliary reward-model training.
In contrast, EG-GRPO uses task-grounded evidence claims annotated in synthetic training data as the supervision units, rewarding newly acquired claims and representing their cumulative set as an explicit knowledge state for state-conditioned advantage estimation. This provides evidence-grounded process supervision without additional human process annotations or auxiliary reward-model training.

\section{Conclusion}
\label{sec:conclusion}
In this paper, we introduced EG-GRPO, a novel reinforcement learning framework that addresses the challenge of sparse rewards in training deep search agents by enabling step-level credit assignment. By leveraging task-grounded evidence claims available in synthetic training data, our method derives dense, task-grounded process rewards without requiring additional human process annotations. To ensure robust and fair advantage estimation, EG-GRPO aligns intermediate actions based on cumulative knowledge states and applies a hierarchical fallback normalization strategy. Extensive evaluations across multiple deep search benchmarks demonstrate that EG-GRPO consistently improves over standard trajectory-level RL and achieves strong performance against existing fine-grained baselines across different model scales. Ultimately, this step-level supervision effectively guides agents to sustain multi-turn exploration and systematically acquire the evidence needed to solve complex search tasks, paving the way for more capable and reliable long-horizon search models.

\section*{AI Use Statement}
\label{sec:ai_use}
In this work, we used generative AI tools for LLM-based judging and annotation within the research pipeline, including final-answer correctness scoring, step-level evidence verification, evidence-claim matching, identification of the first supporting citation for each claim, and automated construction of synthetic evidence claims. Additionally, we used generative AI tools for code development assistance and language editing during manuscript preparation. We reviewed all AI-assisted work, manually verified the reliability of the LLM judge on sampled trajectories, and checked all generated data, annotations, judge outputs, code, and manuscript text. The authors designed the research and experiments, validated and quality-controlled the generated data and annotations, verified implementation correctness and experimental results, and interpreted the findings, and we take full responsibility for the final content of this work.

\section*{Reproducibility Statement}
\label{sec:reproduce}
We provide sufficient detail to reproduce the proposed EG-GRPO training and evaluation pipeline. Section \ref{sec:method} specifies evidence-claim construction, evidence-to-trajectory attribution, incremental step rewards, cumulative knowledge states, three-level fallback grouping, and the final training objective. Section \ref{sec:exp} describes the tool server, baselines, benchmarks, model scales, cold-start SFT and RL training setup, evaluation protocol, main results, analyses, and ablation studies. Appendix \ref{app:training} reports the full SFT and RL configurations, including model initialization, optimization hyperparameters, hardware, batch sizes, context lengths, and EG-GRPO hyperparameters (\(w=0.5\), \(\gamma=0.95\), \(C=5\)). Appendix \ref{app:human_verification} documents human verification of the LLM judge, and Appendix \ref{app:prompt} provides the prompts used for outcome-reward scoring and evidence matching. We provide anonymized implementation code and scripts for reproducing the main experiments in the supplementary material.

\bibliography{iclr2027_conference}
\bibliographystyle{iclr2027_conference}

\clearpage
\appendix
%\section{Appendix}
\label{sec:appendix}
\section{Trajectory Format}
\label{app:format}
Our tool descriptions and trajectory format are shown as follows:

\begin{formatbox}[title=Tool Definitions]
\small
[\\
\hspace*{1em}\{\\
\hspace*{2em}``name'': ``search'',\\
\hspace*{2em}``description'': ``Performs web searches; the tool retrieves the top (default 10) results for each query in one call.'',\\
\hspace*{2em}``parameters'': \{\\
\hspace*{3em}``type'': ``object'',\\
\hspace*{3em}``properties'': \{\\
\hspace*{4em}``query'': \{``type'': ``array'', ``items'': \{``type'': ``string''\},     ``description'': ``Array of query strings. ''\},\\
\hspace*{4em}``num'': \{``type'': ``integer'', ``description'': ``Number of results to return''\},\\
\hspace*{3em}\},\\
\hspace*{3em}``required'': [``query'']\\
\hspace*{2em}\}\\
\hspace*{1em}\},\\
\hspace*{1em}\{\\
\hspace*{2em}``name'': ``visit'',\\
\hspace*{2em}``description'': ``Visit webpages and return the summary of the content.'',\\
\hspace*{2em}``parameters'': \{\\
\hspace*{3em}``type'': ``object'',\\
\hspace*{3em}``properties'': \{\\
\hspace*{4em}``url'': \{``type'': ``array'', ``items'': \{``type'': ``string''\}, ``description'': ``The URL array of the webpages to visit.''\},\\
\hspace*{4em}``goal'': \{``type'': ``string'', ``description'': ``The goal of the visit for webpages''\},\\
\hspace*{3em}\},\\
\hspace*{3em}``required'': [``url'', ``goal'']\\
\hspace*{2em}\}\\
\hspace*{1em}\}\\
]
\end{formatbox}
\captionof{figure}{Tool definitions for the search and visit tools.}
\label{fig:tool-definitions}

\begin{formatbox}[title=Format]
\small
\textcolor{blue}{\texttt{\textless think\textgreater}} thinking process here \textcolor{blue}{\texttt{\textless /think\textgreater}}\\
\textcolor{cyan!70!blue}{\texttt{\textless tool\_call\textgreater}}\\
\{``name'': ``tool name'', ``arguments'': \{``parameter name'': parameter value, ``another parameter name'': another parameter value, ...\}\}\\
\textcolor{cyan!70!blue}{\texttt{\textless /tool\_call\textgreater}}\\
\textcolor{orange}{\texttt{\textless tool\_response\textgreater}}\\
tool\_response here\\
\textcolor{orange}{\texttt{\textless /tool\_response\textgreater}}\\
(more thinking processes, tool calls and tool responses here)\\
\textcolor{blue}{\texttt{\textless think\textgreater}} thinking process here \textcolor{blue}{\texttt{\textless /think\textgreater}}\\
\textcolor{red}{\texttt{\textless answer\textgreater}} answer here \textcolor{red}{\texttt{\textless /answer\textgreater}}
\end{formatbox}
\captionof{figure}{Multi-turn trajectory format for training and inference.}
\label{fig:prompt-format}

\section{Referred Deep Search Agents}
\label{app:reference}

For advance agents with proprietary data, we present the scores of GLM-4.7 ~\citep{zeng2025glm}, Kimi-K2.5~\citep{team2026kimi}, DeepSeek-V3.2~\citep{liu2025deepseek}, Gemini 3.1 Pro~\citep{deepmind_gemini_3_1_pro}, Tongyi-DeepResearch~\citep{team2025tongyi} from their technical reports. For agents with open-source data, we present the score of DeepDive-32B~\citep{lu2025deepdiveadvancingdeepsearch}, WebSailor-32B~\citep{li2025websailornavigatingsuperhumanreasoning}, WebExplorer-8B~\citep{liu2025webexplorerexploreevolvetraining}, OpenSeeker-30B~\citep{du2026openseekerdemocratizingfrontiersearch}, REDSearcher-30B-A3B~\citep{chu2026redsearcherscalablecostefficientframework} from their previous paper.

\section{Data Synthesis Details}
\label{app:data_synthesis}

\subsection{Synthetic QA Construction}
We construct synthetic deep-search queries from structured factual paths derived from an entity-centric knowledge source. Starting from a seed entity, we traverse a multi-hop factual path to obtain a sequence of related entities and factual relations. The resulting path is then transformed into a natural-language question through entity obfuscation, where intermediate entities that should be recovered through search are replaced by descriptive clues while preserving the underlying factual dependencies.

This construction process produces questions that require the agent to progressively recover hidden information through multi-step web search rather than directly matching an entity mentioned in the query. The path length and the number of hidden entities are used to control the reasoning complexity of the synthesized questions.

\subsection{Evidence Claim Extraction}
The structured factual paths used for question synthesis are further converted into atomic evidence claims. For each synthesized query $q$, we obtain a set

$$
\mathcal{C}_q=\{c_1,\ldots,c_K\},
$$

where each $c_k$ is a self-contained factual assertion that can be independently verified from web evidence. Claims are normalized to a consistent granularity so that each claim represents a single verifiable fact required for resolving the query.

For example, for the query: 
\newline
\newline
\textit{A European tour that concluded in a year when a film about a U.S. president was released, and which visited a continent where a writer published research on African literature containing the word 'polyphony' - what was the academic journal where that research appeared?}
\newline
\newline
The original structured facts can be converted into claims such as:
\begin{itemize}
\item $c_1$: The Carpe Diem Tour concluded in 2012.
\item $c_2$: The film \textit{Lincoln} was released in 2012.
\item $c_3$: The Carpe Diem Tour visited Europe.
\item $c_4$: The Carpe Diem Tour visited Africa.
\item $c_5$: Ken Walibora's research was related to Africa.
\item $c_6$: Ken Walibora published research in \textit{Research in African Literatures}.
\item $c_7$: The relevant research article contains ``polyphony'' in its title.
\end{itemize}

These claims constitute the task-grounded evidence units used by EG-GRPO. Although the claims originate from the structured synthesis process, their internal dependency relations are not used by the training algorithm.

\subsection{Claim Verification and Filtering}
To ensure that the extracted evidence claims provide reliable supervision, we apply automatic verification and filtering to the synthesized data. Each claim is required to satisfy three criteria: (1) it represents a concrete factual assertion rather than a vague description; (2) it is independently verifiable from publicly available web evidence; and (3) it provides information that is relevant to resolving the corresponding query.

We further remove samples containing redundant, ambiguous, or unverifiable claims. Claims that require substantial inference beyond the retrieved evidence are also excluded. The resulting claim set is used as the task-grounded evidence annotation for subsequent RL training.

\section{Training Configuration}
\label{app:training}

\paragraph{SFT.} We filtered 1,000 high-quality SFT trajectories from the OpenSeeker~\citep{du2026openseekerdemocratizingfrontiersearch} dataset and adapted them to match the trajectory format used in this paper. Using this dataset, we fine-tuned both the 4B and 30B models for 1 epoch. The training was conducted on 8$\times$ NVIDIA H800 GPUs with a batch size of 32, a learning rate of $2 \times 10^{-5}$, and a maximum context length of 128k.

\paragraph{RL.}Starting from the SFT checkpoints, we conducted RL training on 2,316 synthetic data across 16$\times$ NVIDIA H800 GPUs for 2 epochs, based on verl framework~\citep{jiang2025verltool}. The training used a learning rate of $1 \times 10^{-6}$ and a maximum context length of 64k. For experience collection, we set the prompt batch size to 32 and sampled 8 completions per prompt, yielding a global batch size of 256. Generation parameters during rollout were set to a temperature of 1.0, $\text{top\_p} = 0.95$, $\text{top\_k} = 20$, and a presence penalty of 1.1. For the EG-GRPO advantage calculation, we set the step-level advantage weight $w = 0.5$, discount factor $\gamma = 0.95$, and clipping threshold $C = 5$. We employed Qwen-Flash~\citep{yang2025qwen3} as the judge LLM to perform both final-answer reward scoring and step-level evidence verification.

\section{Human Verification of LLM Judge}
\label{app:human_verification}

To assess the reliability of the LLM judge in identifying evidence acquisition within the EG-GRPO framework, we conducted a manual review of its judgments across 10 randomly sampled deep search trajectories. The sampled trajectories cover 67 evidence claims. Using human assessments as the gold standard, annotators independently verified whether each evidence claim was sufficiently supported by the retrieved information and, when supported, identified the first citation that provided sufficient evidence. We then compared the human annotations with the corresponding LLM judge outputs. The LLM judge achieves 91.0\% accuracy in identifying whether evidence claims are supported and 94.0\% accuracy in identifying the first supporting citation. These results indicate that the LLM judge can reliably identify evidence acquisition and its corresponding supporting citations in search trajectories.

\section{Prompts}
\label{app:prompt}

\begin{formatbox}[title=Prompt for Outcome Rewards]
\small
Judge whether the following \textbf{[response]} to \textbf{[question]} is correct or not based on the precise and unambiguous \textbf{[correct\_answer]} below.\\

\textbf{[question]}: \{question\}\\

\textbf{[response]}: \{response\}\\

Your judgement must be in the format and criteria specified below:\\

\textbf{extracted\_final\_answer}: The final exact answer extracted from the \textbf{[response]}. Put the extracted answer as `None' if there is no exact, final answer to extract from the response.\\

\textbf{[correct\_answer]}: \{correct\_answer\}\\

\textbf{reasoning}: Explain why the \textbf{extracted\_final\_answer} is correct or incorrect based on \textbf{[correct\_answer]}, focusing only on if there are meaningful differences between \textbf{[correct\_answer]} and the \textbf{extracted\_final\_answer}. Do not comment on any background to the problem, do not attempt to solve the problem, do not argue for any answer different than \textbf{[correct\_answer]}, focus only on whether the answers match.\\

\textbf{correct}: Answer `yes' if \textbf{extracted\_final\_answer} matches the \textbf{[correct\_answer]} given above, or is within a small margin of error for numerical problems. Answer `no' otherwise, i.e. if there if there is any inconsistency, ambiguity, non-equivalency, or if the extracted answer is incorrect.
\end{formatbox}
\captionof{figure}{Prompt for outcome rewards.}
\label{fig:judge-format}

\begin{formatbox}[title=Evidence Matching Prompt]
\small
You are an evidence matching expert. Given a list of evidence claims and citation content from a search trajectory, determine which citations first contain each piece of evidence.\\[1em]

\#\# Task\\[0.5em]

For each evidence claim, find the \textbf{earliest} citation number that explicitly contains the stated fact.\\
- Only count citations that explicitly contain the fact; the model's own inferences don't count.\\
- If multiple citations contain the same evidence, only take the smallest number (earliest discovery).\\
- If no citation contains a piece of evidence, its citation\_ids should be an empty list.\\[1em]

\#\# Output Format\\[0.5em]

\#\#\# Matching Analysis of E1\\
Explanation: \{\{Clearly explain whether the citation content explicitly contains the fact stated in E1, citing relevant parts\}\}\\
Evidence Index: \{\{List the citation numbers that contain this evidence\}\}\\
Found: \{\{yes/no\}\}\\[1em]

\#\#\# Matching Analysis of E2\\
Explanation: \{\{Clearly explain whether the citation content explicitly contains the fact stated in E2, citing relevant parts\}\}\\
Evidence Index: \{\{List the citation numbers that contain this evidence\}\}\\
Found: \{\{yes/no\}\}\\[1em]

...\\[1em]

\#\#\# Final JSON-format Summary\\
\{\{\\
  "E1": [3],\\
  "E2": [5],\\
  "E3": []\\
\}\}\\[1em]

Key is the evidence ID, value is a list of supporting citation numbers (starting from earliest discovery).\\[1em]

---\\[1em]

\textbf{[Begin of Evidence List]}\\
\texttt{\{claims\_text\}}\\
\textbf{[End of Evidence List]}\\[1em]

\textbf{[Begin of Citation Content]}\\
\texttt{\{citation\_content\}}\\
\textbf{[End of Citation Content]}
\end{formatbox}

\end{document}